\documentclass[graybox]{svmult}

\usepackage{mathptmx}       
\usepackage{helvet}         
\usepackage{courier}        
\usepackage{type1cm}        
\usepackage{makeidx}         
\usepackage{graphicx}        
\usepackage{multicol}        
\usepackage[bottom]{footmisc}
\usepackage{url}

\usepackage{ifthen}

\begin{document}


\title{Introduction to Presentation Attack Detection\\ in Face Biometrics and Recent Advances}
{
\title*{Introduction to Presentation Attack Detection\\ in Face Biometrics and Recent Advances}
}

\titlerunning{Introduction to Face Presentation Attack Detection} 

\author{Javier Hernandez-Ortega, Julian Fierrez, Aythami Morales and Javier Galbally}



\institute{Javier Hernandez-Ortega \at School of Engineering, Universidad Autonoma de Madrid, Spain, \email{javier.hernandezo@uam.es}
\and Julian Fierrez \at School of Engineering, Universidad Autonoma de Madrid, Spain, \email{julian.fierrez@uam.es}
\and Aythami Morales \at School of Engineering, Universidad Autonoma de Madrid, Spain, \email{aythami.morales@uam.es}
\and Javier Galbally \at European Commission - DG Joint Research Centre, \email{javier.galbally@ec.europa.eu}}

%

%
\maketitle

%


\abstract*{
The main scope of this chapter is to serve as an introduction to face
presentation attack detection, including key resources and advances in the field in the last few years. The next pages present the different
presentation attacks that a face recognition system can confront, in which an
attacker presents to the sensor, mainly a camera, a Presentation Attack Instrument (PAI), that is generally a photograph, a video, or a mask, to try to impersonate a genuine user.
\newline\indent First, we make an introduction of the current status of face
recognition, its level of deployment, and its challenges. In addition,
we present the vulnerabilities and the possible attacks that a face recognition system may be exposed to, showing that way the high importance of presentation
attack detection methods.
\newline\indent We review different types of presentation attack methods, from
simpler to more complex ones, and in which cases they could be effective.
Then, we summarize the most popular presentation attack detection methods to
deal with these attacks.
\newline\indent Finally, we introduce public datasets used by the research
community for exploring vulnerabilities of face biometrics to presentation attacks and developing
effective countermeasures against known PAIs.
}

\abstract{
The main scope of this chapter is to serve as an introduction to face
presentation attack detection, including key resources and advances in the field in the last few years. The next pages present the different
presentation attacks that a face recognition system can confront, in which an
attacker presents to the sensor, mainly a camera, a Presentation Attack Instrument (PAI), that is generally a photograph, a video, or a mask, to try to impersonate a genuine user.
\newline\indent First, we make an introduction of the current status of face
recognition, its level of deployment, and its challenges. In addition,
we present the vulnerabilities and the possible attacks that a face recognition system may be exposed to, showing that way the high importance of presentation
attack detection methods.
\newline\indent We review different types of presentation attack methods, from
simpler to more complex ones, and in which cases they could be effective.
Then, we summarize the most popular presentation attack detection methods to
deal with these attacks.
\newline\indent Finally, we introduce public datasets used by the research
community for exploring vulnerabilities of face biometrics to presentation attacks and developing
effective countermeasures against known PAIs.
}

\section{Introduction}
\label{sec:introduction}

Over the last decades there have been numerous technological advances that helped to bring new possibilities to people in the form of new devices and services. Some years ago, it would have been almost impossible to imagine having in the market devices like current smartphones and laptops, at affordable prices, that allow a high percentage of the population to have their own piece of top-level technology at home, a privilege that historically has been restricted to big companies and research groups.

Thanks to this quick advance in technology, specially in computer science and electronics, it has been possible to broadly deploy biometric systems for the first time. Nowadays, they are present in a high number of scenarios like: border access control~\cite{galbally2019study}, surveillance~\cite{tome2014soft}, smartphone authentication~\cite{alonso2019super}, forensics~\cite{tome2015facial}, and on-line services like e-commerce and e-learning~\cite{hernandez2019edbb}.  

Among all the existing biometric characteristics, face recognition is currently one of the most extended. Face has been studied as a mean of recognition since the 60s, acquiring special relevance in the 90s following the evolution of computer vision \cite{turk1991face}. Some interesting properties of the interaction of human faces with biometric systems are: acquisition at a distance, non-intrusive, and the highly discriminant features of the face to perform identity recognition. 

At present, face\index{Face} is one of the biometric characteristics with the highest economical and social impact due to several reasons:

\begin{itemize}
\item{Face is one of the most largely deployed biometric modes at world level in terms of market quota~\cite{international2007biometrics}. Each day more and more manufacturers are including Face Recognition Systems (FRSs) in their products, like Apple with its Face ID technology. The public sector is also starting to use face recognition for a wide range of purposes like demographic analysis, identification, and access control~\cite{gmf2021}.}
\item{Face is adopted in most identification documents such as the ICAO-compliant biometric passport \cite{gipp2007epassport} or national ID cards \cite{cospedal2004utilizacion}.}
\end{itemize}

Given their high level of deployment, attacks having a FRS as their target are not restricted anymore to theoretical scenarios, becoming a real threat. There are all kinds of applications and sensitive information that can be menaced by attackers. Providing each face recognition application with an appropriate level of security, as it is being done with other biometric characteristics, like iris or fingerprint, should be a top priority.

Historically, the main focus of research in face recognition has been given to the improvement of the performance at verification and identification tasks, that means, distinguishing better between subjects using the available information of their faces. To achieve that goal, a FRS should be able to optimize the differences between the facial features of each user and also the similarities among samples of the same user~\cite{jain2011handbook,tistarelli2017handbook}. Within the variability factors that can affect the performance of face recognition systems there are occlusions, low-resolution, different viewpoints, lighting, etc. Improving the performance of recognition systems in the presence of these variability factors is currently an active and challenging area in face recognition research~\cite{zhao2018towards,li2019low,2018_TIFS_SoftWildAnno_Sosa}.

Contrary to the optimization of their performance, the security vulnerabilities of face recognition systems have been much less studied in the past, and only over the recent few years some attention has been given to detecting different types of attacks~\cite{hadid2015biometrics,li2018face,Galbally2007_Vulnerabilities}. 

Presentation Attacks (PA) can be defined as the presentation of human characteristics or artifacts directly to the input sensor of a biometric system, trying to interfere its normal operation. This category of attacks are highly present in real world applications of biometrics since the attackers do not need to have access to the internal modules of the recognition system. For example, presenting a high quality printed face of a legitimate user to a camera can be enough to compromise a face recognition system if it does not implement proper countermeasures against these artifacts.

The target of face Presentation Attack Detection (PAD) systems is the automated determination of presentation attacks. Face PAD methods aim to distinguish between a legitimate face and a Presentation Attack Instrument (PAI) that tries to mimic bona fide biometric traits. For example, a subset of PAD methods, referred to as liveness detection, involve measurement and analysis of anatomical characteristics or of involuntary and voluntary reactions, in order to determine if a biometric sample is being captured from a living subject actually present at the point of capture. By applying liveness detection, a face recognition system can become resilient against many presentation attacks like the printed photo attacks mentioned previously.

The rest of this chapter is organized as follows: Section~\ref{sec:vulnerability} overviews the main vulnerabilities of face recognition systems, making a description of several presentation attack approaches. Section~\ref{sec:pad_detection} introduces presentation attack detection techniques. Section~\ref{sec:databases} presents some available public databases for research and evaluation of face presentation attack detection. Section~\ref{sec:integration} discusses different architectures and applications of face PAD. Finally, concluding remarks and future lines of work in face PAD are drawn in Section~\ref{sec:conclusion}.

\section{Vulnerabilities in Face Biometrics}
\label{sec:vulnerability}

\begin{figure}[b]
\sidecaption
\includegraphics[scale=.45]{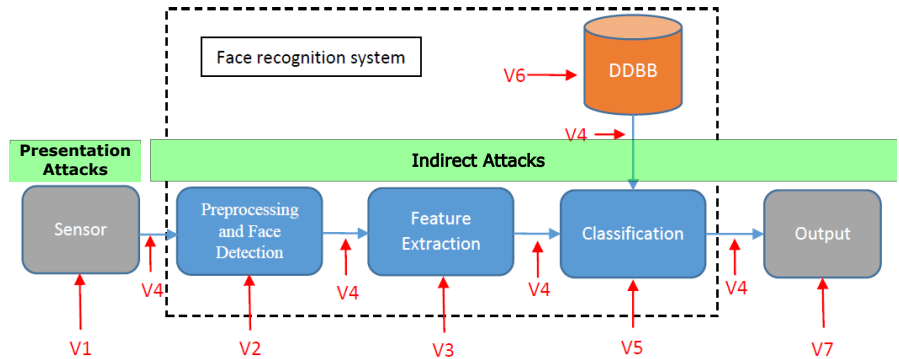}
\caption{\textbf{Scheme of a generic biometric system.} In this type of system, there exist several modules and points that can be the target of an attack (Vulnerabilities V1 to V7). Presentation attacks are performed at sensor level (V1), without the need of having access to the inner modules of the system. Indirect attacks (V2 to V7) can be performed at the databases, the matcher, the communication channels, etc; in this type of attack the attacker needs access to the inner modules of the system and in most cases also specific information about their functioning.}
\label{fig:scheme}     
\end{figure}

In the present chapter we concentrate on Presentation Attacks\index{Presentation Attack (PA)}, i.e. attacks against the sensor of a FRS \cite{galbally2014biometric} (see point V1 in Fig.~\ref{fig:scheme}). Some relevant properties of these attacks are that they require low information about the attacked system and that they present a high success probability when the FRS is not properly protected.

On the contrary, indirect attacks (points V2-V7 in Fig.~\ref{fig:scheme}) are those attacks to the inner modules of the FRS, i.e., the preprocessing module, the feature extractor, the classifier, or the enrolling database. A detailed definition of indirect attacks to face systems can be found in~\cite{gomez2013multimodal}. Indirect attacks can be prevented by improving certain points of the FRS~\cite{martinez11PRLattacksFPs}, like the communication channels, the equipment and infrastructure involved and the perimeter security. The techniques needed for improving those modules are more related to ``classical'' cybersecurity than to biometric techniques. These attacks and their countermeasures are beyond the scope of this book but should not be underestimated.

Presentation attacks are a purely biometric vulnerability that is not shared with other IT security solutions and that needs specific countermeasures. In these attacks, intruders use some type of artefact\index{Artefact}, typically artificial (e.g. a face photo, a mask, a synthetic fingerprint, or a printed iris image), or try to mimic the aspect of genuine users (e.g., gait, signature, or facial expression~\cite{2021_ICIP_EmoVulnerable_Pena}) to present it to the acquisition scanner and fraudulently access the biometric system.  

A high amount of biometric data are exposed, (e.g. photographs and videos at social media sites) showing the face, eyes, voice and behaviour of people. Presentation attackers are aware of this reality and take advantage of those sources of information to try to circumvent face recognition systems \cite{facebooklogin2016}. This is one of the well-known drawbacks of biometrics: ``biometric characteristics are not secrets'' \cite{goodin2008get}. In this context, it is worth noting that the factors that make face an interesting characteristic for person recognition, that is, images that can be taken at a distance and in a non-intrusive way, make it also specially vulnerable to attackers who want to use biometric information in an illicit manner.

In addition to being fairly easy to obtain a face image of the legitimate users, face recognition systems are known to respond weakly to presentation attacks, for example using one of these three categories of attacks: 

\begin{enumerate}
\item{Using a photograph of the user to be impersonated \cite{tan2010face}.}
\item{Using a video of the user to be impersonated \cite{Chingovska_BIOSIG-2012,li2019celebdf}.}
\item{Building and using a 3D model of the attacked face, for example an hyperrealistic mask\index{Mask} \cite{erdogmus2014spoofing}.}
\end{enumerate}

The success probability of an attack may vary considerably depending on the characteristics of the FRS, for example if it uses visible light or works in another range of the electromagnetic spectrum (e.g., infra-red lighting), if it has one or several sensors (e.g., 3D sensors, thermal sensors), the resolution, the lighting; and also depending on the characteristics of the PAI: quality of the texture, the appearance, the resolution of the presentation device, the type of support used to present the fake, or the background conditions.

Without implementing presentation attack detection measures most of the state-of-the-art facial biometric systems are vulnerable to simple attacks that a regular person would detect easily. This is the case, for example, of trying to impersonate a subject using a photograph of his face. Therefore, in order to design a secure FRS in a real scenario, for instance for replacing password-based authentication, Presentation Attack Detection (PAD) techniques should be a top priority from the initial planning of the system.

Given the discussion above, it could be stated that face recognition systems without PAD techniques are at clear risk, so a question often rises: What technique(s) should be adopted to secure them? The fact is that counterfeiting this type of threats is not a straightforward problem, as new specific countermeasures need to be developed and adopted whenever a new attack appears.

With the scope of encouraging and boosting the research in presentation attack detection techniques in face biometrics, there are numerous and very diverse initiatives in the form of dedicated tracks, sessions and workshops in biometric-specific and general signal processing conferences \cite{icassp2017procs,ijcb2017procs}; organization of competitions \cite{chakka2011,chingovska20132nd,boulkenafet2017competition}; and acquisition of benchmark datasets \cite{erdogmus2014spoofing, zhang2012face,george2019biometric,li2019celebdf,livdetface2021} that have resulted in the proposal of new presentation attack detection methods \cite{galbally2014biometric, CVPRw2018}; standards in the area \cite{ISOIEC197922009, ISOIEC3010712016}; and patented PAD mechanisms for face recognition systems \cite{kim2011spoof,raguin2020system}.

\subsection{Presentation Attack Methods}
\label{subsec:methods}

\begin{figure}[b]
\sidecaption
\includegraphics[scale=.50]{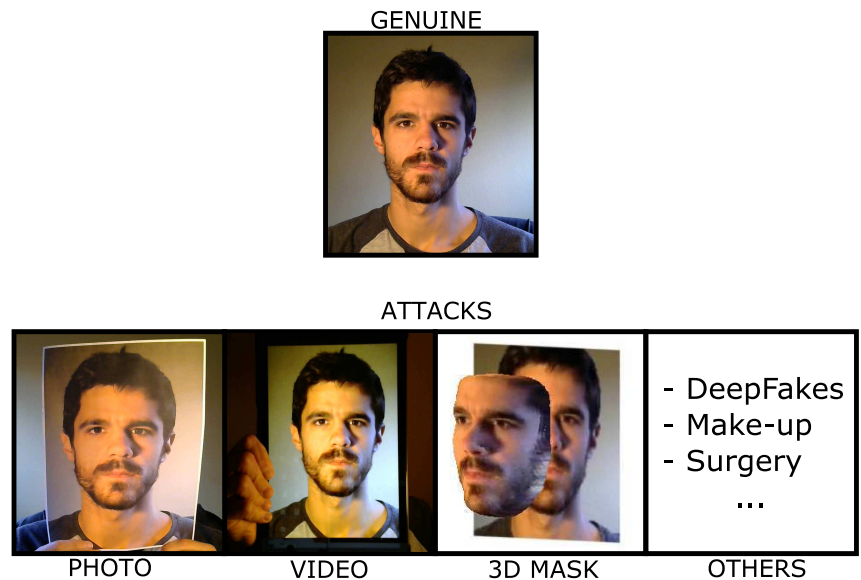}
\caption{\textbf{Examples of face presentation attacks:} The upper image shows an example of a genuine user, and below it there are some examples of presentation attacks, depending of the PAI shown to the sensor: a photo,  a video, a 3D mask, DeepFakes, make-up, surgery, and others.}
\label{fig:attacks}     
\end{figure}

Typically, a FRS can be spoofed by presenting to the sensor (e.g. a camera) a photograph, a video, or a 3D mask\index{3D Mask} of a targeted person (see Fig.~\ref{fig:attacks}). There are other possibilities in order to circumvent a FRS, such as using makeup \cite{dantcheva2012can,dantcheva2017} or plastic surgery. However, using photographs and videos are the most common type of attacks due to the high exposition of face (e.g. social media, video-surveillance), and the low cost of high resolution digital cameras, printers or digital screens.

%

Regarding the attack types, a general classification can be done taking into account the nature and the level of complexity of the PAI used to attack: photo-based, video-based, and mask-based (as can be seen in Fig.~\ref{fig:attacks}). It must be remarked that this is only a classification of the most common types of attacks, but there are some complex attacks that may not fall specifically into in any of these categories, or that may belong to several categories at the same time. This is the case of DeepFake methods, that are usually defined as techniques able to create fake videos by swapping the face of a person by the face of another person, and that could be classified into photo attacks, video attacks, or even mask attacks. In this chapter we have classified those more complex and special attacks in a category named ``Other attacks''.

\subsubsection{Photo Attacks}

%


A photo-attack\index{Photo Attack} consists in displaying a photograph of the attacked identity to the sensor of the face recognition system \cite{anjos2013motion,peng2018face} (see example in Fig.~\ref{fig:attacks}).

Photo attacks are the most critical type of attack to be protected from due to several factors. On the one hand, printing color images from the face of the genuine user is really cheap and easy to do. These are usually called print-attacks in the literature \cite{anjos2011counter}. Alternatively, the photos can be displayed in the high-resolution screen of a device (e.g. a smartphone, a tablet or a laptop \cite{anjos2013motion,Chingovska_BIOSIG-2012,zhang2012face}). On the other hand, it is also easy to obtain samples of genuine faces thanks to the recent growth of social media sites like Facebook, Twitter, and Instagram \cite{facebooklogin2016}. Additionally, with the price and size reduction experimented by digital cameras in recent years, it is also possible to obtain high-quality photos of a legitimate user simply by using a hidden camera.

Among the photo attack techniques there are also more complex approaches like photographic masks. This technique consists in printing a photograph of the subject's face and then making holes for the eyes and the mouth \cite{zhang2012face}. This is a good way to avoid presentation attack detection techniques based on blink detection and in eyes and mouth movement tracking~\cite{shen2021iritrack}.

Even if these attacks may seem too simple to work in a real scenario, some studies indicate that many state-of-the-art systems are vulnerable to them \cite{nguyen2009your,scherhag2017vulnerability,ramachandra2019custom}. Due to their simplicity, implementing effective countermeasures that perform well against them should be a must for any facial recognition system.

\begin{itemize}

\item Information needed to perform the attack: image of the face of the subject to be impersonated.

\item Generation and acquisition of the PAIs: there are plenty of options to obtain high quality face images of the users to be impersonated, e.g., social networks, internet profiles, and hidden cameras. Then, those photographs can be printed or displayed on a screen in order to present them to the sensor of the FRS. 

\item Expected impact of the attack: most basic face recognition systems are vulnerable to this type of attack if specific countermeasures are not implemented. However, the literature offers a large number of approaches with good detection rates of printed photo attacks~\cite{anjos2013motion,bharadwaj2013computationally}.

\end{itemize}

\subsubsection{Video Attacks}

Similarly to the case of photo attacks, video acquisition of people intended to be impersonated is also becoming increasingly easier with the growth of public video sharing sites and social networks, or even using a hidden camera. Another reason to use this type of attack is that it increases the probability of success by introducing liveness appearance to the displayed fake biometric sample \cite{da2012video,pinto2020leveraging}.

Once a video of the legitimate user is obtained, one attacker could play it in any device that reproduces video (smartphone, tablet, laptop, etc) and then present it to the sensor/camera \cite{kim2011motion}, (see Fig. \ref{fig:attacks}). This type of attacks is often referred to in the literature as replay attacks, a more sophisticated version of the simple photo attacks.

Replay attacks\index{Video Replay Attack} are more difficult to detect, compared to photo attacks, as not only the face texture and shape is emulated but also its dynamics, like eye-blinking, mouth and/or facial movements \cite{Chingovska_BIOSIG-2012}. Due to their higher sophistication, it is reasonable to assume that systems that are vulnerable to photo attacks will perform even worse with respect to video attacks, and also that being resilient against photo attacks does not mean to be equally strong against video attacks \cite{zhang2012face}. Therefore, specific countermeasures need to be developed and implemented, e.g., an authentication protocol based on challenge-response \cite{shen2021iritrack}.

\begin{itemize}

\item Information needed to perform the attack: video of the face of the subject to be impersonated.

\item Generation and acquisition of the PAIs: similarly to the case of photo attacks, obtaining face videos of the users to be impersonated is relatively easy thanks to the growth of video sharing platforms (YouTube, Twitch) and social networks (Facebook, Instagram), and also using hidden cameras. The videos are then displayed on a screen in order to present them to the sensor of the FRS.

\item Expected impact of the attack: like in the case of photo attacks most face recognition systems are inherently vulnerable to these attacks, and countermeasures based on challenge-response or in the appearance of the faces are normally implemented. With these countermeasures, classic video attacks have a low success rate.

\end{itemize}

\subsubsection{Mask Attacks}

In this type of attack the PAI is a 3D mask of the user's face\index{Mask Attack}. The attacker builds a 3D reconstruction of the face and presents it to the sensor/camera. Mask attacks require more skills to be well executed than the previous attacks, and also access to extra information in order to construct a realistic mask of the genuine user \cite{liu20163d,CASMAD2018}.

\begin{figure}[b]
\centering
\sidecaption
\includegraphics[scale=.4]{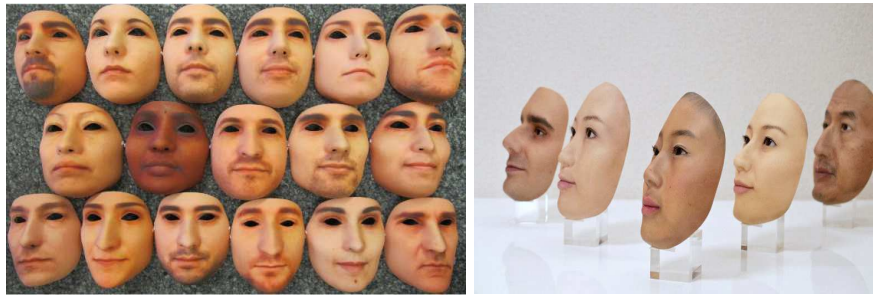}
\caption{\textbf{Examples of 3D masks.} Left) The 17 hard-resin facial masks used to create the 3DMAD dataset, from \cite{erdogmus2014spoofing}. Right) Face Masks built by Real-F reproducing even the eyeballs and finest skin details.}
\label{fig:3DMAD}       
\end{figure}

There are different types of masks depending on the complexity of the manufacturing process and the amount of data that is required. Some examples, ordered from simpler to more complex are:

\begin{itemize}

\item{The simplest method is to print a 2D photograph of the user's face and then stick it to a deformable structure. Examples of this type of structures could be a t-shirt or a plastic bag. Finally, the attacker can put the bag on his face and present it to the biometric sensor \cite{zhang2012face}. This attack can mimic some deformable patterns of the human face, allowing to spoof some low level 3D face recognition systems.}

\item{Image reconstruction techniques can generate 3D models from 2 or more pictures of the genuine user's face, e.g. one frontal photo and a profile photo. Using these photographs, the attacker could be able to extrapolate a 3D reconstruction of the real face\footnote{https://3dthis.com, https://www.reallusion.com/character-creator/headshot} (see Fig. \ref{fig:attacks}). This method is unlikely to spoof top-level 3D face recognition systems, but it can be an easy and cheap option to spoof a high number of standard systems.}

\item{A more sophisticated method consists in making directly a 3D capture of a genuine user's face \cite{galbally2016three,erdogmus2014spoofing,CASMAD2018} (see Fig. \ref{fig:3DMAD}). This method entails a higher level of difficulty than the previous ones since a 3D acquisition can be done only with dedicated equipment and it is complex to obtain without the cooperation of the end-user. However, this is becoming more feasible and easier each day with the new generation of affordable 3D acquisition sensors \cite{realsense2017}.}

\end{itemize}

When using any of the two last methods, the attacker would be able to build a 3D mask with the model he has computed. Even though the price of 3D printing devices is decreasing, 3D printers with suficient quality and definition are still expensive. See references \cite{galbally2016three,CASMAD2018} for examples of 3D-printed masks. There are some companies where such 3D face models may be obtained for a reasonable price\footnote{http://real-f.jp, https://shapify.me, and http://www.sculpteo.com}.


This type of attack may be more likely to succeed due to the high realism of the spoofs. As the complete structure of the face is imitated, it becomes difficult to find effective countermeasures. For example, the use of depth information becomes inefficient against this particular threat. 

These attacks are far less common than the previous two categories because of the difficulties mentioned above to generate the spoofs. Despite the technical complexity, mask attacks have started to be systematically studied thanks to the acquisition of the first specific databases which include masks of different materials and sizes~\cite{erdogmus2014spoofing, galbally2016three, kose2013vulnerability, liu20163d, CASMAD2018,liu2020temporal}.

\begin{itemize}

\item Information needed to perform the attack: 2D masks can be created using only one face image of the user to be impersonated. However, 3D realistic masks usually need $3$ or more face images acquired from different angles.

\item Generation and acquisition of the PAIs: compared to photo and video attacks it is more difficult to generate realistic 3D masks since the attacker needs images of high quality captured from different and complementary angles. Using the photographs, face masks can be orderer to third companies for a reasonable price. 

\item Expected impact of the attack: these attacks are more challenging than photo and video attacks because of the higher realism of the PAIs. However, they are less common due to the difficulty in generating the masks.

\end{itemize}

\subsubsection{Other Attacks}

There are other possibilities in order to circumvent a face recognition system, such as using DeepFake techniques, facial makeup, or modifications via plastic surgery.

Together to the recent availability of large-scale face databases, the progress of deep learning methods like Generative Adversarial Networks (GANs)~\cite{goodfellow2014generative} has led to the emergence of new realistic face manipulation techniques that can be used to spoof face recognition systems. The term Identity Swap (commonly known as DeepFakes) englobes the manipulation techniques consisting in replacing the face of one person in a video with the face of another person. User-friendly applications like FaceApp allow to create fake face images and videos without the need of any previous coding experience. Public information from social networks can be used to create realistic fake videos capable to spoof a FRS, e.g., by means of a replay attack. Recent examples of DeepFake video databases are Celeb-DF~\cite{li2019celebdf} and DFDC~\cite{dolhansky2020deepfake}. DeepFake techniques are evolving very fast, with their outputs becoming more and more realistic each day, so countermeasuring them is a very challenging problem~\cite{tolosana2020deepfakes,hernandez2020deepfakeson}.

Works like~\cite{dantcheva2012can} studied the impact of facial makeup in the accuracy of automatic face recognition systems. They focused their work on determining if facial makeup can affect the matching accuracy of a FRS. They acquired two different databases of females with and without makeup and they tested the accuracy of several face recognition systems whe using those images. They concluded that the accuracy of the FRS is hugely impacted by the presence of facial makeup. 

Nowadays, the technology advancements and the social acceptance have led to a higher presence of plastic surgery among the population. Therefore, the authors of~\cite{singh2010plastic} and~\cite{aggarwal2012sparse} focused their research on determining the level of affectance of face recognition accuracy when using images with presence of plastic surgery that modifies facial appearance. 
In~\cite{singh2010plastic} they reported a high reduction of face recognition accuracy (around a 30\%) of several matching algorithms when comparing images with and without plastic surgery modifications.

\begin{itemize}

\item Information needed to perform the attack: in the case of DeepFake attacks, PAIs can be created only with a few photographs of the targeted subject. Makeup and surgery also need information about the face of the user to be impersonated.

\item Generation and acquisition of the PAIs: similarly to the case of the previous types of attacks, obtaining face images and videos of the users to be impersonated can be easy thanks to video sharing platforms and social networks. Then, the DeepFake videos can be displayed on a screen to present them to the camera of the FRS. Makeup-based attacks need of certain skills to achieve a high quality makeup. Attacks based on surgery modifications are much harder to achieve since they need of highly qualified professionals and of some recovery time after the modifications. 

\item Expected impact of the attack: DeepFake attacks are more difficult to detect than other attacks and how to prevent them is a very active research area nowadays. Surgery attacks are also very difficult to detect since they are actually faces and not synthetic fakes. Makeup attacks, on the other hand, can be detected more easily using PAD techniques based on texture or color, like in the case of photo and video attacks.

\end{itemize}

\section{Presentation Attack Detection}
\label{sec:pad_detection}

\begin{figure}[b]
\centering
\sidecaption
\includegraphics[scale=.23]{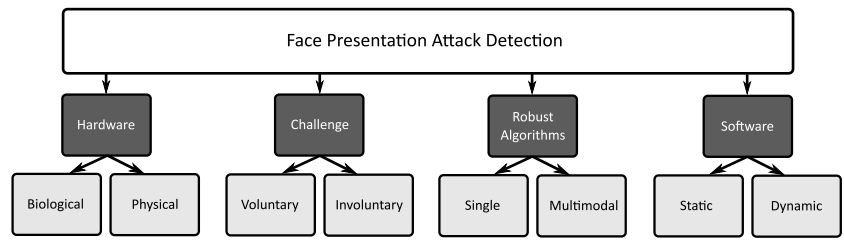}
\caption{Taxonomy of Face Presentation Attack Detection methods.}
\label{fig:taxonomy_PAD}
\end{figure}

Face recognition systems are designed to differentiate between genuine users, not to determine if the biometric sample presented to the sensor is legitimate or a fake. A presentation attack detection method is usually accepted to be any technique that is able to automatically distinguish between legitimate biometric characteristics presented to the sensor and artificially produced PAIs.

Presentation attack detection can be done in four different ways~\cite{hadid2015biometrics}: ($i$) with dedicated hardware to detect an evidence of liveness, which is not always possible to deploy, ($ii$) with a challenge-response method where a presentation attack can be detected by requesting the user to interact with the system in a specific way, ($iii$) employing recognition algorithms intrinsically resilient against attacks, and ($iv$) with software that uses already available sensors to detect any pattern characteristic of live traits. Fig.~\ref{fig:taxonomy_PAD} shows how PAD methods are organized according this proposed taxonomy.

\begin{itemize}

\item \textbf{Hardware-based:} PAD approaches based on dedicated hardware usually take benefit of special sensors like Near Infrared (NIR) cameras~\cite{CVPRw2018}, thermal sensors~\cite{bhattacharjee2018spoofing}, Light Field Cameras (LFC)~\cite{raghavendra2015presentation}, multispectral sensors~\cite{yi2014face}, and 3D cameras~\cite{lagorio2013liveness}. Using the unique properties of the different types of dedicated hardware, the biological and physical characteristics of legitimate faces can be measured and distinguished from PAIs more easily, e.g., processing temperature information with thermal cameras~\cite{sun2011tir} and estimating the 3D volume of the artifacts thanks to 3D acquisition sensors~\cite{lagorio2013liveness}. However, these approaches are not popular even though they tend to achieve high presentation detection rates, because in most systems the required hardware is expensive and not broadly available.

\item \textbf{Challenge-Response:} These PAD methods present a challenge to the user, i.e., completing a predefined task, or expose them to a stimulus in order to record their voluntary or involuntary response. Then that response is analyzed to decide if the access attempt comes from a legitimate user or from an attacker. An example of this approach can be found in~\cite{sluganovic2016using}, where the authors studied the involuntary eye response of users exposed to a visual stimulus. Other examples are~\cite{shen2021iritrack} where the authors requested the users to make specific movements with their eyes and~\cite{chou2021presentation} where the users were indicated to say some predefined words. Challenge-response methods can be effective against many presentation attacks but they usually require more time and cooperation from users' side, something that is not always possible or desirable.

\item \textbf{Robust algorithms:} Existing face recognition systems can be designed or trained to learn how to distinguish between legitimate faces and PAIs, making them inherently robust to some types of presentation attacks. However, developing face recognition algorithms intrinsically robust against presentation attacks is not straightforward and the most common approach consists in relying on  multimodal biometrics to increment the security level thanks to the information coming from the other biometric modalities.


\end{itemize}

The limitations of these three types of PAD methods, together with the easiness of deploying software-based PAD methods, have made most of the literature on face Presentation Attack Methods (PAD) to be focused on running software-based PAD algorithms over already deployed hardware. This is why in the next part of the chapter we focus on describing software-based PAD and the different categories that compose this type of approach. 

\subsection{Software-based face PAD}


\begin{table}[t]
\scriptsize
\centering
\caption{\textbf{Selection of relevant works in software-based face PAD.}}
\label{tab:PAD_works}       
%
%
\resizebox{1\textwidth}{!}{
\begin{tabular}{|c|c|c|c|c|c|}
\hline\noalign{\smallskip}
Method & Year & Type of Images & Database used & Type of features \\
\noalign{\smallskip}\svhline\noalign{\smallskip}
\cite{kim2009masked} & 2009 & Visible and IR photo & Private & Color (reflectance) - Hand Crafted\\
\cite{anjos2011counter} & 2011 & RGB video & PRINT-ATTACK & Face-background motion - Hand Crafted\\
\cite{Chingovska_BIOSIG-2012} & 2012 & RGB video & REPLAY-ATTACK  & Texture based - Hand Crafted\\
\cite{yang2013face} & 2013 & RGB photo and video & NUAA PI, PRINT-ATTACK and CASIA FAS & Texture based - Hand Crafted\\
\cite{bharadwaj2013computationally} & 2013 & RGB photo and video & PRINT-ATTACK and REPLAY ATTACK & Texture based - Hand Crafted\\
\cite{anjos2013motion} & 2013 & RGB video & PHOTO-ATTACK & Motion correlation analysis - Hand Crafted\\
\cite{galbally2014image} & 2014 & RGB video & REPLAY-ATTACK & Image Quality based - Hand Crafted\\
\cite{smith2015face} & 2015 & RGB video & Private & Color (challenge reflections) - Hand Crafted\\
\cite{li2016generalized} & 2016 & RGB video & 3DMAD and private & rPPG (color based) - Hand Crafted\\
\cite{OULUNPU} & 2017 & RGB video & OULU-NPU & Texture based - Hand Crafted\\
\cite{CVPRw2018} & 2018 & RGB and NIR video & 3DMAD and private & rPPG (color based) - Hand Crafted\\
\cite{george2019biometric} & 2019 & RGB, Depth and NIR video & WMCA & Fine-tuned face recog. features - Deep Learning\\
\cite{hernandez2020deepfakeson} & 2020 & RGB video & Celeb-DF v2 and DFDC & rPPG (color based) - Deep Learning\\
\cite{george2020learning} & 2021 & RGB, Depth, Thermal, and NIR video & WMCA, MLFP, and SiW-M & Spoof-specific info. - Deep Learning\\
\cite{yu2021transrppg} & 2021 & RGB video & 3DMAD and HKBU-MARsV2 & rPPG (color based) - Deep Learning\\

\noalign{\smallskip}\hline\noalign{\smallskip}
\end{tabular}
}
\end{table}
Software-based PAD methods are convinient in most of the cases since they allow to upgrade the countermeasures in existing systems without needing new pieces of hardware and permitting authentication to be done in real time without extra user interaction. Table~\ref{tab:PAD_works} shows a selection of relevant PAD works based on software techniques, including information about the type of images they use, the databases in which they are evaluated, and the types of features they analyze. The table also illustrates the current dominance of deep learning among PAD methods, since like in many other research areas, during the last years most state-of-the-art face PAD works have changed from methods based on hand-crafted features to deep learning approaches based on architectures like Convolutional Neural Networks and Generative Adversarial Networks. 

Regardless of whether they belong to one category (hand-crafted) or the other (deep learning), software-based PAD methods can be divided into two main categories depending on whether they take into account temporal information or not: static\index{Static PAD Methods} and dynamic analysis\index{Dynamic PAD Methods}. 

\subsubsection{Static Analysis}

This subsection refers to the development of techniques that analyze static features like the facial texture to discover unnatural characteristics that may be related to presentation attacks.

The key idea of the texture-based approach is to learn and detect the structure of facial micro-textures that characterise real faces but not fake ones. Micro-texture analysis has been effectively used in detecting photo attacks from single face images: extraction of texture descriptions such as Local Binary Patterns (LBP) \cite{Chingovska_BIOSIG-2012} or Gray-Level Co-occurrence Matrices (GLCM) followed by a learning stage to perform discrimination between textures.

For example, the recapturing process by a potential attacker, the printing of an image to create a spoof, usually introduces quality degradation in the sample, making it possible to distinguish between a genuine access attempt and an attack, by analyzing their textures \cite{galbally2014image}. 

The major drawback of texture-based presentation attack detection is that high resolution images are required in order to extract the fine details from the faces that are needed for discriminating genuine faces from presentation attacks. These countermeasures will not work properly with bad illumination conditions that make the captured images to have bad quality in general.

Most of the time, the differences between genuine faces and artificial materials can be seen in images acquired in the visual spectrum with or without a preprocessing stage. However, sometimes a translation to a more proper feature space \cite{zhang2015pca}, or working with images from outside the visible spectrum \cite{gonzalez2017exploring} is needed in order to distinguish between real faces and spoof-attack images. 

Aditionally to the texture, there are other properties of the human face and skin that can be exploited to differentiate between real and fake samples. Some of these properties are: absorption, reflection, scattering, and refraction \cite{kim2009masked}.

This type of approaches may be useful to detect photo-attacks, video-attacks, and also mask-attacks, since all kinds of spoofs may present texture or optical properties different than real faces.

In recent years, to improve the accuracy of traditional static face PAD methods the researchers have been focused on applying the power of deep learning to face PAD mainly using transfer-learning from face recognition. This technique makes possible to adapt facial features learned for face recognition to face presentation attack detection without the need of a huge amount of labeled data. This is the case of~\cite{george2019biometric} where the authors transfer facial features learned for face recognition and used them for detecting presentation attacks. Finally, to increase the generalization ability of their method to unseen attacks, they fused the decisions of different models trained with distinct types of attacks. 


However, even though deep learning methods have shown to be really accurate when evaluated in intra-database scenarios, their performance usually drops significantly when they are tested under inter-database scenarios. Deep learning models are capable of learning directly from data, but they are normally overfitted to the training databases, causing poor generalization of the resulting models when facing data from other sources. To avoid this, the most recent works in the literature face this problem by implementing domain generalization techniques. For example, the authors of~\cite{george2020learning} introduced a novel loss function to force their model to learn a compact embedding for genuine faces while being far from the embeddings of the different presentation attacks.

\subsubsection{Dynamic Analysis}

These techniques have the target of distinguishing presentation attacks from genuine access attempts based on the analysis of motion. The analysis may consist in detecting any physiological sign of life, for example: pulse~\cite{2020_MediComp_HRcomparison_JHO}, eye-blinking~\cite{2020_ICMI_mEBAL_Daza}, facial expression changes~\cite{2021_ICIP_EmoVulnerable_Pena}, or mouth movements. This objective is achieved using knowledge of the human anatomy and physiology. 

As stated in Section~\ref{sec:vulnerability}, photo attacks are not able to reproduce all signs of life because of their static nature. However, video attacks and mask attacks can emulate blinking, mouth movements, etc. Related to these types of presentation attacks, it can be assumed that the movement of the PAIs, differs from the movement of real human faces which are complex nonrigid 3D objects with deformations.

One simple aproximation to this type of countermeasures consists in trying to find correlations between the movement of the face and the movement of the background respect to the camera \cite{anjos2013motion, kim2011motion}. If the fake face presented contains also a piece of fake background, the correlation between the movement of both regions should be high. This could be the case of a replay attack, in which the face is shown on the screen of some device. This correlation in the movements allows to evaluate the degree of synchronization within the scene during a defined period of time. If there is no movement, as in the case of a fixed support attack, or too much movement, as in a hand-based attack, the input data is likely to come from a presentation attack. Genuine authentication will usually have uncorrelated movement between the face and the background, since user's head generally moves independently from the background.

Other example of this type of countermeasures is~\cite{pinto2020leveraging} where the authors propose a method for detecting face presentation attacks based on properties of the scenario and the facial surfaces such as albedo, depth, and reflectance.

A high number of the software-based PAD techniques are based on liveness detection without needing any special help of the user.  These presentation attack detection techniques aim to detect some physiological signs of life such as eye blinking~\cite{2020_ICMI_mEBAL_Daza,yang2013face,pan2008liveness}, facial expression changes~\cite{2021_ICIP_EmoVulnerable_Pena}, and mouth movements.

Other works like \cite{bharadwaj2013computationally} provide more evidence of liveness using Eulerian video magnification \cite{wu2012eulerian} applying it to enhance small changes in face regions, that often go unnoticed. Some changes that are amplified thanks to this technique are, for example, small color and motion changes on the face caused by the human blood flow, by finding peaks in the frequency domain that correspond to the human heartbeat rate. Works like~\cite{CVPRw2018,yu2021transrppg,hernandez2020deepfakeson} use remote photoplethysmography for liveness detection, and more specifically 3D mask PAD, without relying on the appearance features of the spoof like the texture, shape, etc.

As metioned above, motion analysis approaches usually require some level of motion between different head parts or between the head and the background. Sometimes this can be achieved through user cooperation \cite{smith2015face}. Therefore, some of these techniques can only be used in scenarios without time requirements as they may need time for analyzing a piece of video and/or for recording the user's response to a command. Due to the nature of these approaches, some videos and well-performed mask attacks may deceive the countermeasures.

\section{Face Presentation Attacks Databases}
\label{sec:databases}

In this section we overview some publicly available databases\index{PAD Databases} for research in face PAD. The information contained in these datasets can be used for the development and evaluation of new face PAD techniques against presentation attacks.

As it has been mentioned in the past sections, with the recent spread of biometric applications, the threat of presentation attacks has grown, and the biometric community is starting to acquire large and comprehensive databases to make recognition systems more resilient against presentation attacks.

International competitions have played a key role to promote the development of PAD measures. These competitions include the recent LivDet-Face 2021~\cite{livdetface2021}, the CelebA-Spoof Challenge 2020~\cite{li2019celebdf}, the 
ChaLearn Face Antispoofing Attack Detection Challenge 2019~\cite{chalearn2019}, the Multi-modal Face Anti-spoofing (Presentation Attack Detection) Challenge 2019~\cite{zhang2019dataset}, the Competition on Generalized Face Presentation Attack Detection in Mobile Authentication Scenarios 2017 \cite{boulkenafet2017competition}, and the 2011 and 2013 2D Face Anti-Spoofing contests~\cite{chakka2011,chingovska20132nd}.

Despite the increasing interest of the community in studying the vulnerabilities of face recognition systems, the availability of PAD databases is still scarce. The acquisition of new datasets is highly difficult because of two main reasons:

\begin{itemize}

\item{Technical aspects: the acquisition of presentation attack data offers additional challenges to the usual difficulties encountered in the acquisition of standard biometric databases \cite{ortega2010multiscenario} in order to correctly capture similar fake data than the present in real attacks (e.g., generation of multiple types of PAIs).}

\item{Legal aspects: as in the face recognition field in general, data protection limits the distribution or sharing of biometric databases among research groups. These legal restrictions have forced most laboratories or companies working in the field of presentation attacks to acquire their own datasets usually small and limited.}

\end{itemize}

In the area of face recognition PAD, we can find the following public databases (ordered chronologically following their publication date):

\begin{itemize}


\item{The NUAA Photo Imposter Database (NUAA PI DB) \cite{tan2010face} was one of the first efforts to generate a large public face PAD dataset. It contains images of real access attempts and print-attacks of $15$ users. The images contain frontal faces with a neutral expression captured using a webcam. Users were also told to avoid eye-blinks. The attacks are performed using printed photographs on photographic paper. Examples from this database can be seen in Fig.~\ref{fig:NUAA_DB}. The NUAA PI DB is property of the Nanjing University of Aeronautics and Astronautics.}\\

\begin{figure}[b]
\sidecaption
\includegraphics[scale=.25]{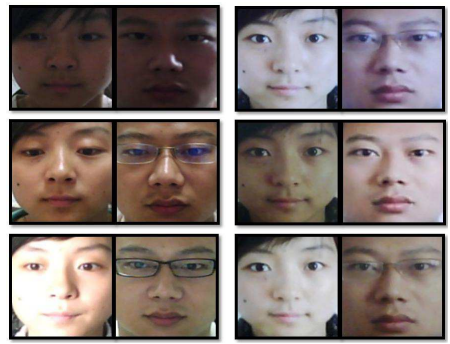}
\caption{\textbf{Samples from the NUAA Photo Imposter Database \cite{tan2010face}.} Samples from two different users are shown. Each row corresponds to one different session. In each row,
the left pair are from a live human and the right pair from a photo fake. Images have been taken from \cite{tan2010face}.}
\label{fig:NUAA_DB}       
\end{figure}



\item{The PRINT-ATTACK DB \cite{anjos2011counter} represents another step in the evolution of face PAD databases, both in terms of the size ($50$ different users were captured) and of the types of data acquired (it contains video sequences instead of still images). It only considers the case of photo attacks. It consists of $200$ videos of real accesses and $200$ videos of
print attack attempts from $50$ different users. Videos were recorded under two different background and illumination conditions. Attacks were carried out with hard copies of high resolution photographs of the $50$ users, printed on plain A4 paper. The PRINT-ATTACK DB is property of the Idiap Research Institute.}\\



\item{The REPLAY-ATTACK database \cite{Chingovska_BIOSIG-2012}, is an extension of the PRINT-ATTACK database. It contains short videos of both real-access and presentation attack attempts of $50$ different subjects. The attack attempts present in the database are $1300$ photo and video attacks using mobile phones and tablets under different lighting conditions. The attack attempts are also distinguished depending on how the attack device is hold: hand-based and fixed-support. Examples from this database can be seen in Fig.~\ref{fig:REPLAY_DB}.}\\

\begin{figure}[b]
\sidecaption
\includegraphics[scale=.3]{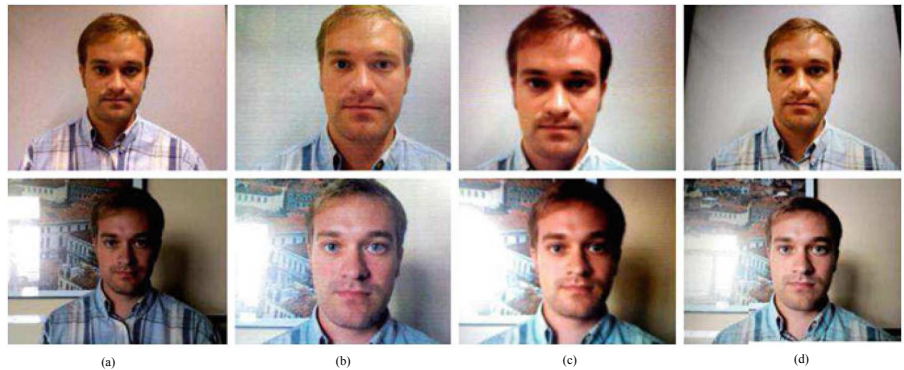}
%
%
\caption{\textbf{Examples of real and fake samples from the REPLAY-ATTACK DB \cite{Chingovska_BIOSIG-2012}.} The images come from videos acquired in two illumination and background scenarios (controlled and adverse). The first row belongs to the controlled scenario while the second row represents the adverse conditions. $(a)$ Shows real samples, $(b)$ shows samples of a printed photo attack, $(c)$ corresponds to a LCD photo attack, and $(d)$ to a high-definition photo attack.}
\label{fig:REPLAY_DB}       
\end{figure}



\item{The 3D MASK-ATTACK DB (3DMAD) \cite{erdogmus2014spoofing}, as its name indicates, contains information related to mask-attacks. As described above, all previous databases contain attacks performed with 2D spoofing artifacts (i.e., photo or video) that are very rarely effective against systems capturing 3D face data. It contains access attempts of $17$ different users. The attacks were performed with real-size 3D masks manufactured by ThatsMyFace.com\footnote{http://www.thatsmyface.com/}. For each access attempt a video was captured using the Microsoft Kinect for Xbox 360, that provides RGB data and also depth information. That allows to evaluate both 2D and 3D PAD techniques, and also their fusion \cite{galbally2016three}. Example masks from this database can be seen in Fig.~\ref{fig:3DMAD}. 3DMAD is property of the Idiap Research Institute.}\\



\item{The OULU-NPU DB \cite{OULUNPU}, contains information of presentation attacks acquired with mobile devices. Nowadays mobile authentication is one of the most relevant scenarios due to the wide spread of the use of smartphones. However, in most datasets the images are acquired in constrained conditions. This type of data may present motion, blur, and changing illumination conditions, backgrounds and head poses. The database consists of $5,940$ videos of $55$ subjects recorded in three distinct illumination conditions, with $6$ different smartphone models. The resolution of all videos is 1920$\times$1080 and it comprehends print and video-replay attacks. The OULU-NPU DB is property of the University of Oulu and it has been used in the IJCB 2017 Competition on Generalized Face Presentation Attack Detection \cite{boulkenafet2017competition}. Examples from this database can be seen in Fig.~\ref{fig:oulu}.}\\

\begin{figure}[b]
\centering
\includegraphics[width=0.85\linewidth]{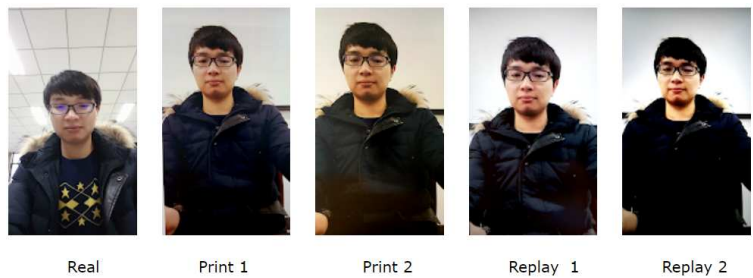}
%
%
\caption{\textbf{Examples of bonafide and attack samples from OULU-NPU DB \cite{OULUNPU}.} The images come from videos acquired with mobile devices. The figure shows a legitimate sample, two examples of print attacks and other two examples of replay attacks. Image extracted from https://sites.google.com/site/oulunpudatabase.}
\label{fig:oulu}       
\end{figure}



\item{The Custom Silicone Mask Attack Dataset (CSMAD)~\cite{CASMAD2018} (collected at the Idiap Research Institute) contains 3D mask presentation attacks. It is comprised of face data from $14$ subjects, from which $6$ subjects has been selected to construct realistic silicone masks (made by Nimba Creations Ltd.). The database contains $87$ bonafide videos and $159$ attack videos captured under four different lighting conditions. CSMAD is composed of RGB, Depth, NIR, and thermal videos of $10$ seconds of duration.}\\


\item{The Wide Multi Channel Presentation Attack Database (WMCA)~\cite{george2019biometric} consists of video recordings of $10$ seconds for both bonafide attemps and presentation attacks. The database is composed of RGB, Depth, NIR, and thermal videos of $72$ different subjects, with $7$ sessions for each subject. WCMA contains 2D and 3D presentation attacks including print, replay, and paper and silicone mask attacks. The total number of videos in the database is $1,679$ ($1,332$ are attacks).}\\ 



\item{The CelebA-Spoof Database~\cite{li2019celebdf} is a large-scale face anti-spoofing dataset with $625,537$ images from $10,177$ different subjects. The database includes labels for $43$ attributes related to the face, the illumination conditions, the environment, and the spoof types. The spoof images were captured on $2$ different environments and under $4$ illumination conditions using $10$ different acquisition devices. CelebA-Spoof contains $10$ different attack types, e.g., print attacks, replay attacks, and 3D masks.}\\


\item{The LiveDet Database~\cite{livdetface2021}, is a dataset built as a combination of data from two of the organizers of the Face Liveness Detection Competition (LivDet-Face 2021). The final database contains data from $48$ subjects, with a total of $724$ images and $814$ videos (of $6$ seconds) acquired using $5$ different sensors including reflex cameras and mobile devices. $8$ different presentation artifacts were used for the images and $9$ for the videos, comprehending paper photographs, photograps shown from digital screens (e.g., a laptop), paper masks, 3D silicone masks, and video replays.}\\

%

\end{itemize}

Finally, we include the description of two of the most challenging DeepFake databases up to date~\cite{2020_JSTSP_GANprintR_Neves,2021_ICPR_DeepFakesRegions_Tolosana}, i.e., Celeb-DF v2 and DFDC. The videos in these databases contain DeepFakes with a large range of variations in face sizes, illumination, environments, pose variations, etc. DeepFake videos can be used for example, to create a video with the face of a user and use it for a replay attack against a FRS.

\begin{itemize}

\item{Celeb-DF v2}~\cite{li2019celebdf} is a database that consists of $590$ legitimate videos extracted from Youtube, corresponding to celebrities of different gender, age, and ethnic group. Regarding fake videos, a total of $5$,$639$ videos were created swapping faces using DeepFake technology. The average length of the face videos is around $13$ seconds (at $30$ fps).\\

\item{DFDC Preview Database}~\cite{dolhansky2019deepfake} is one of the latest public databases, released by Facebook in collaboration with other institutions like Amazon, Microsoft, and the MIT. The DFDC Preview dataset consists of $1$,$131$ legitimate videos from $66$ different actors, ensuring realistic variability in gender, skin tone, and age. A total of $4$,$119$ videos were created using two different DeepFake generation methods by swapping subjects with similar appearances.\\

\end{itemize}

In Table~\ref{tab:databases} we show a comparison of the most relevant features of all the databases described in this section.

\begin{table}[t]
\scriptsize
\centering
\caption{\textbf{Features of the main public databases for research in face PAD.} Comparison of the most relevant features of each of the databases described in this chapter.}
\label{tab:databases}       
%
%
\resizebox{1\textwidth}{!}{
\begin{tabular}{|c|c|c|c|c|c|}
\hline\noalign{\smallskip}
Database & Users \# (real/fakes) & Samples \# (real/fakes) & Attack Types & Support & Attack Illumination \\
\noalign{\smallskip}\svhline\noalign{\smallskip}
NUAA PI \cite{tan2010face} & 15/15 & 5,105/7,509 & Photo  & Held & Uncont.\\
REPLAY-ATTACK \cite{anjos2011counter, anjos2013motion, Chingovska_BIOSIG-2012} & 50/50 & 200/1,000 & Photo and Replay & Held and Fixed & Cont. \& Uncont.\\
3DMAD \cite{erdogmus2014spoofing} & 17/17 & 170/85 & Mask  & Held & Cont.\\
OULU-NPU \cite{OULUNPU} & 55/55 & 1,980/3,960 & Photo and Replay  & Mobile & Uncont.\\
CSMAD \cite{CASMAD2018} & 14/6 & 87/159 & Photo and Replay  & Held and Fixed & Cont.\\
WMCA \cite{george2019biometric} & 72/72 & 347/1,332 & Photo, Replay, and Mask  & Held & Uncont.\\
CelebA-Spoof \cite{li2019celebdf} & 10,177/10,177 & 202,599/422,938 & Photo, Replay, and Mask  & Held & Uncont.\\
LiveDet \cite{livdetface2021} & 48/48 & 125/689 & Photo, Replay, and Mask  & Held & Uncont.\\
CelebDF-v2 \cite{li2019celebdf} & 59/59 & 590/5,639 & DeepFakes  & - & Uncont.\\
DFDC Preview \cite{dolhansky2019deepfake} & 66/66 & 1,131/4,119 & DeepFakes  & - & Uncont.\\
\noalign{\smallskip}\hline\noalign{\smallskip}
\end{tabular}
}
$^a$ Containing also PHOTO-ATTACK DB and PRINT-ATTACK DB
\end{table}

\section{Integration with Face Recognition Systems}
\label{sec:integration}

In order to create a face recognition system resistant to presentation attacks, the proper PAD techniques have to be selected. After that, the integration of the PAD countermeasures with the FRS can be done at different levels, namely, score-level\index{Score-level fusion} or decision-level fusion\index{Decision-level fusion} \cite{chingovska2013anti,aguilar2006adapted}. 

\begin{figure}[t]
\sidecaption
\includegraphics[scale=.45]{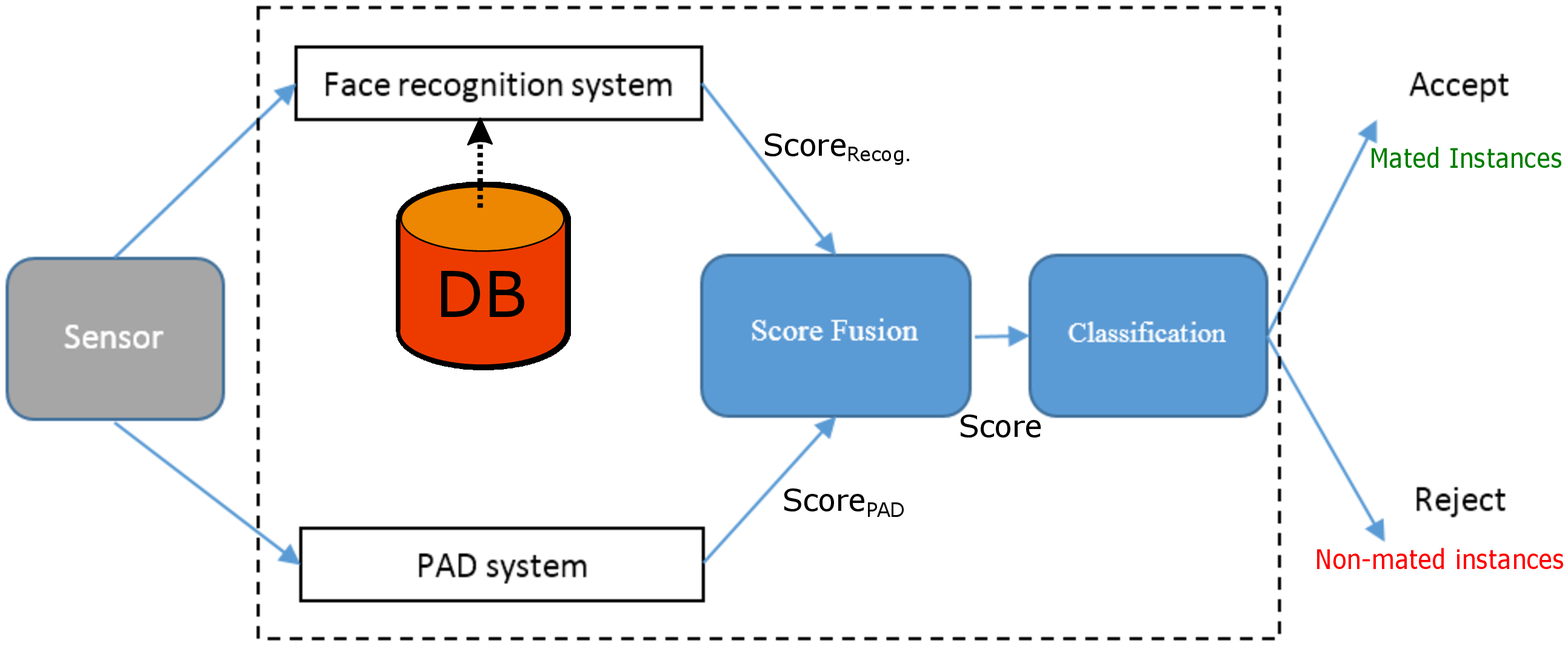}
\caption{\textbf{Scheme of a parallel score-level fusion between a PAD and a face recognition system.} In this type of scheme, the input biometric data is sent at the same time to both the face recognition system and the PAD system, and each one generates a independent score,  then the two scores are fused to take one unique decision.}
\label{fig:parallel_fusion}       
\end{figure}

The first possibility consists in using score level fusion as shown in Fig.~\ref{fig:parallel_fusion}. This is a popular approach due to its simplicity and the good results given in fusion of multimodal biometric systems \cite{de2013can, fierrez2018fusion, ross2006handbook}. In this case, the biometric data enter at the same time to both the face recognition system and the PAD system, and each one computes their own scores. Then the scores from each system are combined into a new final score that is used to determine if the sample comes from a genuine user or not. The main advantage of this approach is its speed, as both modules, i.e. the PAD and face recognition modules, perform their operations at the same time. This fact can be exploited in systems with good parallel computation specifications, such as those with multicore/multithread processors.

\begin{figure}[t]
\sidecaption
\includegraphics[scale=.45]{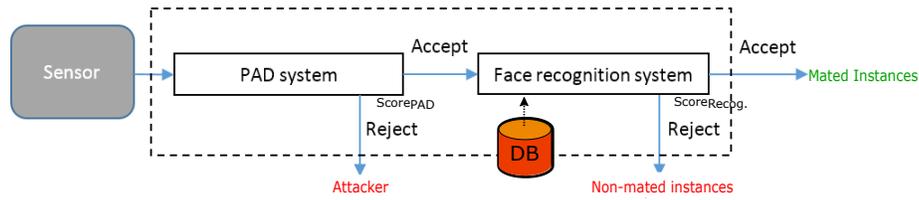}
\caption{\textbf{Scheme of a serial fusion between a PAD and a face recognition system.} In this type of scheme the PAD system makes its decision first, and only if the samples are determined to come from a living person, then they are processed by the face recognition system.}
\label{fig:series_fusion}       
\end{figure}

Another common way to combine PAD and face recognition systems is a serial scheme, as in Fig.~\ref{fig:series_fusion}, in which the PAD system makes its decision first, and only if the samples are determined to come from a living person, then they are processed by the face recognition
system. Thanks to this decision-level fusion, the FRS will search for the identity that corresponds to the biometric sample knowing previously that the sample does not come from a presentation attack. Differently to the parallel approach, in the serial scheme the average time for an access attempt will be longer due to the consecutive delays of the PAD and the face recognition modules. However, this approach avoids extra work to the face recognition system in the case of a PAD attack, since it should be detected in an early stage.

\section{Conclusion and Look Ahead on Face PAD}
\label{sec:conclusion}

Face recognition systems are increasingly being deployed in a diversity of scenarios and applications. Due to this widespread use, they have to withstand a high variety of attacks. Among all these threats, one with high impact are presentation attacks.

In this chapter, an introduction of the strengths and vulnerabilities of face as a biometric characteristic has been presented, including key resources and advances in the field in the last few years. We have described the main presentation attacks, differentiating between multiple approaches, the corresponding PAD countermeasures, and the public databases that can be used to evaluate new protection techniques. The weak points of the existing countermeasures have been stated, and also some possible future directions to deal with those weaknesses have been discussed. 

Due to the nature of face recognition systems, without the correct PAD countermeasures, most of the state-of-the-art systems are vulnerable to attacks since they do not integrate any module to discriminate between legitimate and fake samples. Usually, PAD techniques are developed to fight against one concrete type of attack (e.g. printed photos), retrieved from a specific dataset. The countermeasures are thus designed to achieve high presentation attack detection against that particular spoof technique. However, when testing these same techniques against other type of PAIs (e.g. video-replay), usually the system is unable to efficiently detect them. There is one important lesson to be learned from this fact: there is not a superior PAD technique that outperforms all the others in all conditions; so knowing which technique to use against each type of attack is a key element. It would be interesting to use different countermeasures that have proved to be effective against particular types of PAIs, in order to develop fusion schemes that combine their results, achieving that way a high performance against a variety of presentation attacks data \cite{fierrez2018fusion,de2013can,hadid2015biometrics}. 

In addition, as technology progresses constantly, new hardware devices and software techniques continue to appear. This is the case of the recent and popular DeepFakes, a term that englobes those methods capable of generating images and videos with very realistic face spoofs using deep learning methods and few input data. From the detection point of view it is also important to keep track of this quick technological progress in order to use it to develop more efficient presentation attack detection techniques. For example, using the power of deep learning and some of its associated techniques like transfer-learning has shown to improve the accuracy of PAD methods in the recent years~\cite{george2019biometric}. Additionally, focusing the research on the biological nature of biometric characteristics (e.g. thermogram, blood flow, etc.) should be considered~\cite{hernandez2020deepfakeson,li2016generalized}, as the standard techniques based on texture and movement seem to be inefficient against some PAIs. 

Additionally, it is of the utmost importance to collect new databases with new scenarios in order to develop more effective PAD methods. Otherwise, it will be difficult to grant an acceptable level of security of face recognition systems. However, it is especially challenging to recreate realistic attacking conditions in a laboratory evaluation. Under controlled conditions, systems are tested against a restricted number of typical PAIs. These restrictions make it unfeasible to collect a database with all the different fake spoofs that may be found in the real world.

To conclude this introductory chapter, it could be said that even though a great amount of work has been done to fight face presentation attacks, there are still big challenges to be addressed in this topic, due to the evolving nature of the attacks, and the critical applications in which these systems are deployed in the real world.

\begin{acknowledgement}
This work was mostly done in the context of the TABULA RASA and BEAT projects funded under the 7\textsuperscript{th} Framework Programme of EU. The 3\textsuperscript{rd} Edition update has been made in the context of EU H2020 projects PRIMA and TRESPASS-ETN. This work was also partially supported by the Spanish project BIBECA (RTI2018-101248-B-I00 MINECO/FEDER).
\end{acknowledgement}
%

\bibliographystyle{spmpsci}
\bibliography{references}

\end{document}